\documentclass[journal]{IEEEtranTIE}
\usepackage[utf8]{inputenc}
\usepackage{graphicx}
\usepackage{cite}
\usepackage{picinpar}
\usepackage{amsmath}
\usepackage{url}
\usepackage{flushend}
\usepackage{colortbl}
\usepackage{soul}
\usepackage{multirow}
\usepackage{pifont}
\usepackage{color}
\usepackage{alltt}
\usepackage[hidelinks]{hyperref}
\usepackage{enumerate}
\usepackage{siunitx}
\usepackage{breakurl}
\usepackage{epstopdf}
\usepackage{pbox}
\usepackage{fontawesome5}
\usepackage{booktabs}
\usepackage{hyperref}
\usepackage{marvosym}
\usepackage{etoolbox}
\usepackage{changepage}

\begin{document}
\title{EasyUUV: An LLM-Enhanced Universal and Lightweight Sim-to-Real Reinforcement Learning Framework for UUV Attitude Control}

\author{
	\vskip 1em
	
Guanwen Xie$^{1,}$$^\dagger$, \IEEEmembership{Student Member, IEEE}, Jingzehua Xu$^{2,}$$^{\dagger,}$\textsuperscript{\Letter}, \IEEEmembership{Student Member, IEEE},\\Jiwei Tang$^{3}$, \IEEEmembership{Student Member, IEEE}, Yubo Huang$^{4}$, Zixi Wang$^{5}$, Shuai Zhang$^{6}$, \IEEEmembership{Member, IEEE},\\ Dongfang Ma$^{7}$, \IEEEmembership{Member, IEEE}, Juntian Qu$^{1}$, \IEEEmembership{Member, IEEE}, and Xiaofan Li$^{2}$, \IEEEmembership{Member, IEEE}

\thanks{$^{1}$G. Xie and J. Qu are with Tsinghua Shenzhen International Graduate School, Tsinghua University, Shenzhen, 518055, China. E-mail: xgw24@mails.tsinghua.edu.cn, juntian.qu@sz.tsinghua.edu.cn.}%
\thanks{$^{2}$J. Xu and X. Li are with Department of Mechanical Engineering, The University of Hong Kong, Pokfulam Road, Hong Kong, China; E-mail: xjzh23@berkeley.edu, lixf@hku.hk.}
\thanks{$^{3}$J. Tang is with Department of Data and Systems Engineering, The University of Hong Kong, Pokfulam Road, Hong Kong, China; Email: tangjiwei@connect.hku.hk.}
\thanks{$^{4}$Y. Huang is with School of Civil Engineering, Southwest Jiaotong University, Chengdu, 611756, China; E-mail: ybforever@my.swjtu.edu.cn.}
\thanks{$^{5}$Z. Wang is with School of Information and Software Engineering, University of Electronic Science and Technology of China, Chengdu, 611731, China; E-mail: 202521090118@std.uestc.edu.cn.}
\thanks{$^{6}$S. Zhang is with Department of Data Science, New Jersey Institute of Technology, NJ 07102, USA. E-mail: sz457@njit.edu.}
\thanks{$^{7}$D. Ma is with Ocean College, Zhejiang University, Zhoushan, 316021, China; Email: mdf2004@zju.edu.cn.}
\thanks{$^\dagger$ These authors contributed equally to this work.}
\thanks{$\textsuperscript{\Letter}$ Corresponding author.}
}

\makeatletter
\newcommand{\@colorBib@list}{}%
\newcommand{\addcolorBib}[1]{\listadd{\@colorBib@list}{#1}}
\AtBeginEnvironment{thebibliography}{%
	\let\old@bibitem\bibitem
	\renewcommand{\bibitem}[2][]{%
		\normalcolor
		\if\relax\detokenize{#1}\relax
		\old@bibitem{#2}%
		\else
		\old@bibitem[#1]{#2}%
		\fi
		\ifinlist{#2}{\@colorBib@list}{\color{blue}}{}
	}
}
\makeatother

\maketitle
	
\begin{abstract}
Despite recent advances in Unmanned Underwater Vehicle (UUV) attitude control, existing methods still struggle with generalizability, robustness to real-world disturbances, and efficient deployment. To address the above challenges, this paper presents EasyUUV, a Large Language Model (LLM)-enhanced, platform-agnostic, and lightweight simulation-to-reality reinforcement learning (RL) framework for robust attitude control of UUVs. EasyUUV combines parallelized RL training with a hybrid control architecture, where a learned policy outputs high-level attitude corrections executed by an adaptive S-Surface controller. A multimodal LLM is further integrated to adaptively tune controller parameters at runtime using visual and textual feedback, enabling training-free adaptation to unmodeled dynamics. Also, we have developed a low-cost 6-DoF UUV platform and applied an RL policy trained through efficient parallelized simulation. Extensive simulation and real-world experiments validate the effectiveness and adaptive performance of EasyUUV in achieving robust and adaptive UUV attitude control across diverse underwater conditions. To facilitate reproducibility, the source code, LLM prompts, video, and the supplementary material are provided in the following repositories:

{\faGithub\ Homepage}: \url{https://360zmem.github.io/easyuuv/}

{\faYoutube\ Video}: \ \ \ \ \ \ \ \ \url{https://youtu.be/m2yLQzxiILc}

\begin{adjustwidth}{1.3em}{0pt}

{\faFile \ Supplementary Material}: \ 
\href{https://drive.google.com/file/d/1ImMiyGIPoPyj2ATnXQwbiuOroTSZ2SSs}
{\nolinkurl{https://drive.google.com/file/}}\\
\hspace*{5.8em}
\href{https://drive.google.com/file/d/1ImMiyGIPoPyj2ATnXQwbiuOroTSZ2SSs}
{\nolinkurl{d/1ImMiyGIPoPyj2ATnXQwbiuOroTSZ2SSs}}

\end{adjustwidth}

\end{abstract}
\vspace{-1.3mm}
\begin{IEEEkeywords}
Unmanned Underwater Vehicle, Reinforcement Learning, Large Language Model, Simulation to Reality, Attitude Control
\end{IEEEkeywords}

\markboth{ARXIV preprint}
{}

\definecolor{limegreen}{rgb}{0.2, 0.8, 0.2}
\definecolor{forestgreen}{rgb}{0.13, 0.55, 0.13}
\definecolor{greenhtml}{rgb}{0.0, 0.5, 0.0}

\section{Introduction}
\label{sec:introduction}
\IEEEPARstart{U}{nmanned} Underwater Vehicles (UUVs) are transforming underwater operations, playing critical roles in marine research \cite{1}, environmental monitoring \cite{2}, and resource exploration \cite{3}. However, achieving robust and intelligent autonomy for UUVs, especially in attitude control, remains an open challenge. UUVs operate in complex, highly dynamic, and partially observable environments, where nonlinear hydrodynamics, ocean currents, and wave disturbances introduce significant uncertainty \cite{OysterNet}. These factors complicate the design of reliable attitude control systems, which are essential for high-stakes missions such as coral reef navigation \cite{4}, pipeline inspection \cite{5}, and sample retrieval \cite{6}.

Traditional and mainstream controllers, such as PID \cite{7}, Model Predictive Control (MPC) \cite{8}, Sliding Mode Control (SMC) \cite{7}, and Fuzzy Logic Control (FLC) \cite{9}, provide partial solutions but are often hindered by their reliance on accurate dynamics modeling or their limited adaptability. Their performance often degrades in uncertain conditions due to modeling inaccuracies, hysteresis, and control overshoots, raising risks in unstructured real-world deployments \cite{10,11}.

\begin{figure*}[!t]
\centering
\includegraphics[width=0.979\textwidth]{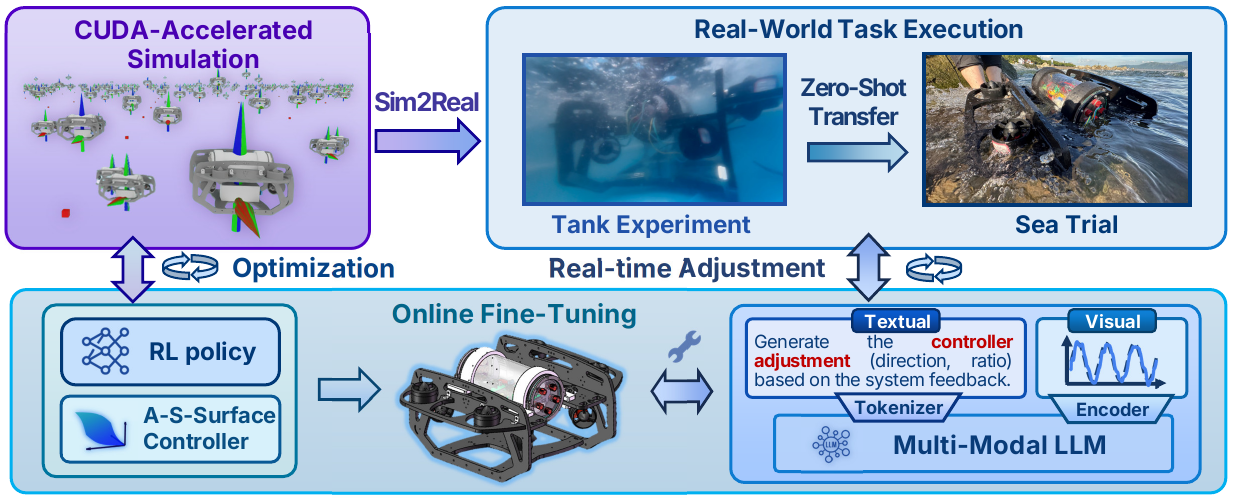}
\vspace{-2mm}
\caption{\textbf{Illustration of our developed EasyUUV framework}.
EasyUUV is an LLM-enhanced platform-agnostic and lightweight Sim2Real RL framework for UUV attitude control, which trains the expert policy via RL in parallelized simulation, while transferring it to a real UUV platform. A multimodal LLM agent further adapts controller parameters using dynamics and sensor feedback for robust performance.}
\label{fig_1}
\vspace{-2mm}
\end{figure*}

Reinforcement Learning (RL), by contrast, has emerged as a promising data-driven alternative for autonomous agents to learn robust control policies through interaction \cite{30}. In particular, unlike traditional controllers that have inadequate adaptivity or depend on accurate system identification, RL can learn directly from experience, thereby eliminating the need for precise hydrodynamic modeling and enabling end-to-end optimization toward task objectives \cite{31}. As a result, RL is especially well-suited for underwater environments, which are nonlinear, partially observable, and difficult to model analytically. Building on this advantage, RL's adaptability to high-dimensional, nonlinear dynamics has motivated extensive UUV-related research \cite{12,13,14}. Nevertheless, despite these strengths, three major challenges remain: the simulation-to-reality (Sim2Real) gap, limited generalizability, and deployment inefficiency. To address these issues, domain randomization can partially mitigate model mismatch by injecting variability into simulation \cite{12}; however, it still cannot fully prevent instability under attitude perturbations or parameter shifts \cite{16}. Moreover, the high computational cost of RL training and the lack of generalizable hydrodynamic/thruster models further constrain its practical application across diverse UUV platforms.

In addition to the challenges described above, the real-world deployment of RL-based controllers often requires extensive manual tuning to handle variations in vehicle dynamics, environmental conditions, and sensor noise, particularly when transitioning across different UUV platforms or operational domains \cite{29}. This lack of adaptability not only hinders scalability but also increases the risk of degraded performance or mission failure in unfamiliar conditions \cite{18}. Fortunately, the introduction of the large language model (LLM) enables online, training-free adaptation of controller parameters \cite{19}. By leveraging historical system trajectories, real-time sensory feedback, and task-specific context encoded in both visual and textual forms, the LLM can dynamically adjust key control parameters without interrupting operations \cite{20}. This capability enhances the robustness and generalizability of the control system, allowing a single RL-trained policy to maintain stable performance across diverse and uncertain underwater environments \cite{21}.

Based on the above analysis, we developed EasyUUV, a lightweight, platform-agnostic Sim2Real RL framework enhanced with LLM (as shown in Fig. 1), where the term \textit{universal} in the title specifically denotes this \textit{platform-agnostic} and \textit{embodiment-agnostic} capability. Different from conventional UUV controllers that mainly rely on accurate dynamic modeling or manual tuning, EasyUUV adopts a role-separated learning-control architecture rather than simply stacking RL, LLM, and control modules. Specifically, the RL policy outputs platform-agnostic high-level attitude/depth correction commands, which are executed by the A-S-Surface controller and actuation layer to preserve structured low-level stability and disturbance rejection. To support efficient training and robust transfer, EasyUUV leverages an Isaac Lab~\cite{25}-based simulation environment with hydrodynamic modeling, parallelized RL training, and domain randomization. During real-world deployment, a multimodal LLM agent further adjusts selected controller parameters using visual and textual feedback, but its outputs are restricted to bounded adjustment directions and scaling ratios rather than arbitrary control commands. In this way, EasyUUV provides a deployment-oriented framework that combines offline Sim2Real robustness, structured low-level control, and online adaptation for real-world UUV attitude control.

Our main contributions are summarized as follows:
\begin{itemize}
    \item \textbf{A platform-agnostic role-separated hybrid control framework for UUV Sim2Real attitude control:} We propose a lightweight modular framework where the RL policy generates attitude/depth corrections, while the A-S-Surface controller and actuation layer map them to platform-specific control signals. This improves adaptability under model mismatch without direct black-box thruster-level control.

    \item \textbf{An efficient parallelized RL training and zero-shot deployment pipeline:} We build an Isaac Lab-based simulation environment with hydrodynamic modeling, thruster modeling, and domain randomization over key physical and control parameters. This enables efficient GPU-accelerated training and improves robustness during zero-shot transfer to the real UUV platform.

    \item \textbf{A constrained multimodal LLM-based online tuning mechanism:} We integrate an off-the-shelf multimodal LLM in a zero-shot, prompt-based manner to adjust selected controller parameters using visual logs, historical tracking responses, and textual feedback. Without task-specific pre-training or fine-tuning, LLM outputs only bounded adjustment directions and scaling ratios, improving runtime adaptability while reducing unsafe updates.

    \item \textbf{Real-world validation on a low-cost 6-DoF UUV platform:} We develop a compact and low-cost UUV platform and validate EasyUUV through simulation, indoor tank experiments, disturbance tests, and sea trials, demonstrating practical Sim2Real feasibility and robust attitude-control performance under real underwater conditions.
\end{itemize}

The remainder of this paper is organized as follows: Section II introduces the architecture and core modules of the proposed EasyUUV framework. Section III describes the experimental setup and presents both simulation and real-world results, along with detailed analysis. Finally, Section IV concludes the paper and discusses limitations and future research directions.

\section{Architecture and Modules}
In this section, we introduce the EasyUUV framework in detail, including both simulation and hardware platforms.
\subsection{Architecture Overview}

\setcounter{figure}{1}
\begin{figure*}[!t]
    \centering
    \includegraphics[width=0.979\linewidth]{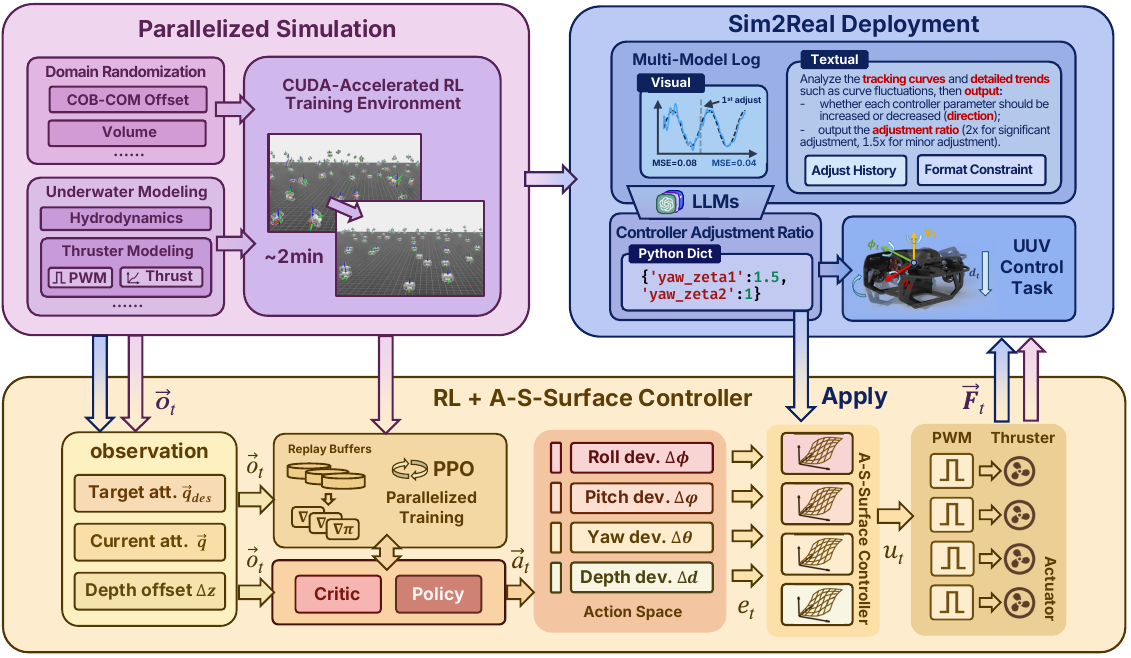}
    \vspace{-2mm}
    \caption{\textbf{Architecture of the EasyUUV framework}, which comprise three parts: (a) RL and A-S-Surface-based composite controller module; (b) Parallelized RL training simulation environment developed on Isaac Lab; and (c) Sim2Real deployment module for real-world adaption.} 
    \label{fig_2}
    \vspace{-2mm}
\end{figure*}

As shown in Fig.~\ref{fig_2}, the proposed EasyUUV framework consists of three tightly coupled components: a composite control architecture, a parallelized RL training environment, and a Sim2Real deployment pipeline. Specifically, the composite controller integrates an RL policy, a nonlinear A-S-Surface controller, and an LLM-based adaptation module, enabling both learning-driven decision-making and structured stability enforcement. The parallelized simulation environment incorporates hydrodynamic effects and thruster models to support efficient data generation and policy training, while the Sim2Real deployment pipeline transfers the trained policy to real UUV platforms and supports online adaptation during operation. The RL policy takes the target attitude, current estimated attitude, and depth offset as observations, and outputs deviation commands that are subsequently transformed by the A-S-Surface controller into low-level control inputs and PWM signals for the thrusters.

This architecture is designed to address the practical difficulties of real-world UUV attitude control while avoiding a fully unconstrained end-to-end control structure. In particular, the RL policy is not used as a direct low-level actuator controller; instead, it outputs bounded high-level attitude and depth correction commands, while the A-S-Surface controller performs real-time low-level execution, disturbance rejection, and thruster-command generation. The RL policy is trained in an NVIDIA Isaac Lab-based simulation environment with MuJoCo-based hydrodynamic modeling~\cite{24}, where domain randomization is applied to key parameters such as COB-COM offsets~\cite{12}, UUV volume, and controller gains to improve robustness across varying vehicle dynamics. After training, the learned policy is directly deployed to the real UUV platform without additional fine-tuning. During deployment, the multimodal LLM further adjusts only selected controller parameters according to historical tracking responses, visual logs, and textual feedback, rather than generating thruster commands or arbitrary control signals. Therefore, the stability and robustness of EasyUUV are supported by a bounded hierarchical structure: the inner-loop A-S-Surface controller maintains real-time attitude stabilization, the RL policy provides bounded reference corrections, and the LLM operates only as an event-triggered high-level tuning assistant. Robustness against physical-parameter mismatch is additionally strengthened during training through the domain randomization scheme of Table~I.

\subsection{Simulation Platform and Controller Design}
A carefully designed simulation platform is essential for efficient RL training, zero-shot policy transfer, and robust real-world deployment. In this work, we develop a simplified yet hardware-agnostic simulation platform integrated with NVIDIA Isaac Lab to support the proposed LLM-enhanced RL-based UUV attitude control framework. Considering the Sim2Real difficulty of underwater control, the simulator is not intended to exactly reproduce every underwater condition, since hydrodynamic forces, buoyancy effects, thruster nonlinearities, COB-COM offsets, and external disturbances are strongly coupled in real deployment. Instead, it provides a physically informed and computationally efficient environment where the RL policy can experience representative dynamic variations before being transferred to the real UUV platform.

For \textbf{hydrodynamic modeling}, we augment each physics step ($120\,\text{Hz}$) with a custom hydrodynamic wrench layer adapted from MuJoCo's phenomenological fluid model \cite{24}. Each body is approximated by an equivalent inertia box with half-dimensions $r_i\!=\!\sqrt{\tfrac{3}{2m}(I_{jj}\!+\!I_{kk}\!-\!I_{ii})}$, used to compute drag and viscous forces/torques in body frame; the resulting wrench is fed back to the simulator through the \texttt{set\_external\_force\_and\_torque} interface.

The previous augmentation enables the calculation of total fluid forces $\mathbf{f}_\text{inertia} \!=\! \mathbf{f}_D \!+\! \mathbf{f}_V$ and torques $\mathbf{g}_\text{inertia} \!=\! \mathbf{g}_D \!+\! \mathbf{g}_V$, incorporating both drag and viscous effects. Drag forces and torques are modeled as $f_{D,i} \!=\! -2\rho r_j r_k |v_i|v_i$ and $g_{D,i} \!=\! -\frac{1}{2}\rho r_i(r_j^4 \!+\! r_k^4)|\omega_i|\omega_i$, while viscous terms are $f_{V,i} \!=\! -6\beta\pi r_\text{eq} v_i$ and $g_{V,i} \!=\! -8\beta\pi r_\text{eq}^3 \omega_i$,\! with $r_\text{eq} \!=\! (r_x \!+\! r_y \!+\! r_z)\!/3$ and $\beta$ denoting the fluid viscosity.

For \textbf{thruster dynamics}, we implement a realistic actuation pipeline that modulates thrust via PWM signals to electronic speed controllers. Based on empirical data from Blue Robotics T200 thrusters at 16V \cite{26}, the thrust output $\tau_\Omega$ (in N) is modeled as a function of normalized input $a \in [-1, 1]$, corresponding to 1100-1900 \textmu s PWM, using a piecewise quadratic fit:
\begin{equation}
\tau_\Omega \!=\!
\begin{cases}
29.54 a^2 + 26.10 a - 2.44, & a \!\in\! (0.08, 1], \\
0, & a \!\in\! [-0.08,\! 0.08], \\
-21.75 a^2 + 21.75 a + 2.07, & a \!\in\! [-1, -0.08).
\end{cases}
\end{equation}

\begin{table}
  \centering
  \caption{Domain Randomization Configuration Details} 
  \vspace{-2mm}
  \label{tab:1}

    \begin{tabular}{lcc}
      \toprule
      \textbf{Parameters} & \textbf{Distributions} & \textbf{Values (Low, High)}\\
      \midrule
        COB-COM offset (m) & Uniform Sphere & ($\pm$0.075, $\pm$0.15)\\
        Volume (L) & Uniform & ($\pm$1.5, $\pm$3)\\
        \multirow{2}{*}{Controller Gain} & \multirow{2}{*}{Uniform} & ($\pm$15, $\pm$30)\% of the\\ 
        & & relative value \\
      \bottomrule 
  \end{tabular}
  \vspace{-5mm}
\end{table}

To improve policy generalization and real-world adaptability, we apply domain randomization within the high-efficiency parallelized simulation environment. During training, key parameters such as the COB-COM offset, volume, and controller gains are randomly perturbed to account for structural and dynamic variations, where the COB-COM offset is particularly important because it affects the gravity- and buoyancy-induced torques in attitude control. All simulation models are implemented in Isaac Lab and trained with GPU acceleration, and the full set of randomized parameters is listed in Table~\ref{tab:1}. This strategy reduces the risk that the learned policy overfits to a single nominal simulated model. Although domain randomization cannot cover all possible underwater conditions, it exposes the policy to a broader range of physical and control variations during training, thereby improving robustness to model mismatch during zero-shot deployment.

Building on this simulation environment, we implement an \textbf{RL policy} using the RSL-RL library \cite{28} with Proximal Policy Optimization (PPO) \cite{PPO} for training. The UUV observes a 9-dimensional state vector $\vec{o}_t = \{\vec{q}, \vec{q}_{\text{des}}, \Delta z\}$, where $\Delta z$ represents the depth error, while $\vec{q}$ and $\vec{q}_{\text{des}}$ denote the current and desired attitude quaternions, which ensure singularity-free orientation tracking. The policy then outputs a 4-dimensional action vector $\vec{a}_t = \{\Delta \phi, \Delta \varphi, \Delta \theta, \Delta d\}$, representing deviations in roll, pitch, yaw, and depth, which are passed to the low-level controller. This four-dimensional action space is adopted because the present work focuses on attitude and depth control rather than full 6-DoF trajectory tracking or navigation. Specifically, roll, pitch, and yaw directly correspond to the three attitude-control objectives, while depth is included to regulate heave-related motion that is strongly coupled with underwater attitude stability. In contrast, surge and sway mainly correspond to horizontal translation and are more relevant to path following, station keeping with horizontal position constraints, or navigation tasks, which are beyond the current scope of this paper. Therefore, excluding surge and sway from the policy output keeps the action space compact, reduces unnecessary exploration, and improves learning efficiency for the attitude/depth stabilization problem considered here. Nevertheless, this does not reduce the physical UUV model to a 4-DoF system: the simulator, hydrodynamic model, hardware platform, and thruster actuation still retain 6-DoF dynamics, while the A-S-Surface controller and actuation layer map the high-level policy commands to executable low-level control signals. In the simulation environment, no additional range constraint is imposed on the attitude-control angles beyond the wrapped angular representation, allowing the controller to be evaluated without the practical safety bounds required by the physical UUV platform. The reward function composed of three terms guides the policy toward stable behavior. The terms are listed as follows:
\begin{itemize}
    \item $r_q = \exp(-|\vec{q} \vec{q}_{\text{des}}^{\star}|)$ encourages orientation alignment,
    \item $r_p = \exp(-||\vec{a}||^b)$ penalizes excessive control actions (with $b=1$),
    \item $r_z = \exp(-||\Delta z||^2)$ promotes accurate depth tracking.
\end{itemize}
\quad These terms are then linearly weighted to guide the policy toward stable and efficient behavior.

At the low level, we employ an \textbf{A-S-Surface controller} \cite{23} to ensure a fast and robust response under underwater disturbances. Based on the system state $\mathbf{x}(t) = [\delta(t), \dot{\delta}(t)]^\top$, the angle error and its derivative are defined as $e(t) = \delta_{\text{des}} - \delta(t)$ and $\dot{e}(t) = -\dot{\delta}(t)$. In implementation, the angular error is normalized into the principal interval to avoid discontinuities at the $\pm180^\circ$ boundary. By defining a sliding-like surface variable $s(t) = \zeta_1 e(t) + \zeta_2 \dot{e}(t)$ (where $\zeta_1 > 0$ and $\zeta_2 > 0$ represent the proportional and derivative-like controller gains, respectively, which serve as the primary parameters tuned by the LLM), the control output is computed as:
\begin{equation}
u_t = \frac{2}{1 + \exp(-s(t))} - 1 + \Delta u(t),
\end{equation}
with the adaptive compensation term updated in discrete time by:
\begin{equation}
\Delta u(t+1) = \Delta u(t) + \alpha s(t),
\end{equation}
where $\alpha$ is a tunable learning rate. This formulation offers both high gain for large deviations and smooth convergence near the setpoint. A rigorous, continuous-time Lyapunov stability analysis (where the continuous update law is $\dot{\Delta u}(t) = \alpha s(t)$) and asymptotic convergence proof via Barbalat's Lemma are provided in Section~I of the supplementary material~\cite{Barbalat}.

To address runtime variations during zero-shot Sim2Real deployment, an asynchronous, cloud-based multimodal LLM (specifically GPT-4o~\cite{GPT}) adaptively tunes the controller parameters without retraining. The LLM performs zero-shot, prompt-based reasoning over operational feedback without task-specific pre-training or fine-tuning. Queried through the OpenAI API, GPT-4o receives the latest visual response log (encoded as a Base64 image) alongside a textual prompt containing state observations, active control axes, and historical parameters and MSE values. The complete visual snapshots and raw system prompts are detailed in Section~IV of the supplementary material. Operating as an asynchronous supervisory layer, the LLM infers parameter adjustments solely from existing deployment feedback, avoiding the instability risks of online active exploration. Specifically, the LLM processes two types of input: 
\begin{itemize}
    \item Visual logs, which provide recent response curves and tuning trends of the enabled attitude axes, allowing the LLM to inspect tracking error, overshoot, oscillation, convergence, and response delay;
    \item Textual and numerical data, including sensor readings, user/task instructions, historical MSE values, previous controller parameters, enabled control axes, and previous tuning records.
\end{itemize}

\setcounter{figure}{2}
\begin{figure}[!t]
    \centering
    \includegraphics[width=0.986\linewidth]{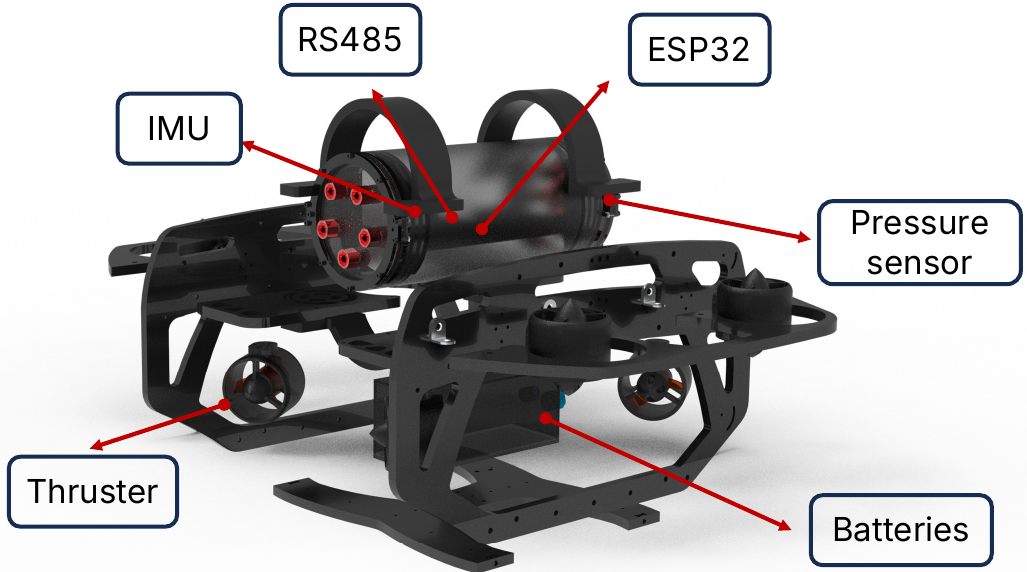}
    \vspace{-2mm}
    \caption{Exploded view of our EasyUUV hardware platform.} %
    \label{fig_3}
    \vspace{-2mm}
\end{figure}

Constructed as a structured tuning assistant, the LLM operates with a clear controller prior: the prompt defines the role of $\zeta_1$ (determining response strength and convergence speed) and $\zeta_2$ (regulating auxiliary overshoot). Contextual inputs couple visual logs with numerical records, enabling the LLM to relate performance trends to previous parameter adjustments. To prevent unconstrained outputs, responses are strictly restricted to a Python dictionary of multiplicative factors (e.g., \texttt{\{'roll\_zeta1':1.05, 'yaw\_zeta1':0.67\}}). These scaling factors are selected from a dynamically partitioned action space containing both coarse candidates (e.g., $2.0, 1.5, 0.67, 0.5$) for transient emergency recovery and fine-grained candidates (e.g., $1.05, 0.95$) for steady-state fine-tuning, as detailed in our multi-scale paradigm and Section~IV of the supplementary material. The prompt also enforces safety rules, such as halting $\zeta_1$ increases when actuator saturation or negligible performance improvement is detected. The LLM executes only for event-triggered, high-level tuning under tracking degradation, rather than in the high-frequency loop. Consequently, the LLM does not generate raw actuator commands, but outputs bounded gain modifications that are applied at a slower timescale. This structured integration improves interpretability, automates parsing, and mitigates unsafe gain updates while preserving low-level parameter stability~\cite{32}. 


\setcounter{figure}{4}
\begin{figure*}[!t]
    \centering
    \includegraphics[width=0.99\linewidth]{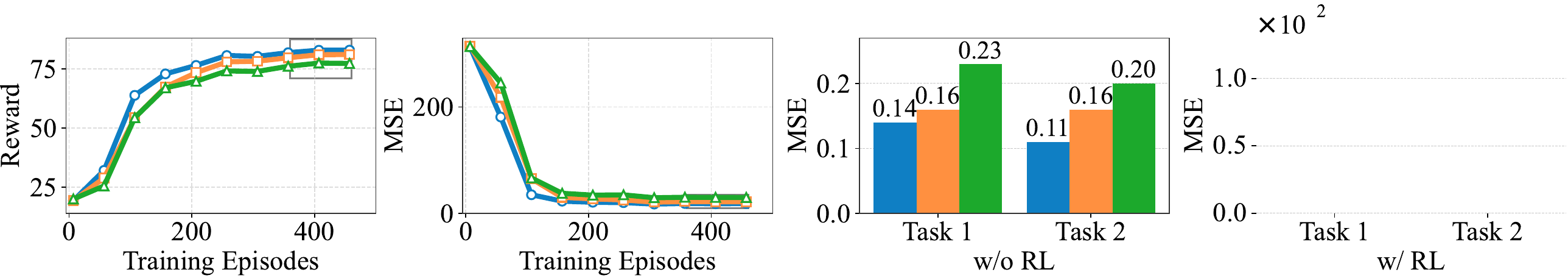}
    \vspace{-2mm}
    \caption{RL training convergence and tracking MSE comparison across different control strategies. (Left): Cumulative reward curves during parallelized training, (Middle): MSE convergence curves during training, and (Right): comparative tracking MSE across Task 1 and Task 2 under both RL-enabled and non-RL configurations.}
    \label{fig_5}
    \vspace{-2mm}
\end{figure*}

\subsection{Hardware Platform}
Our EasyUUV hardware platform (Fig.~3) is a compact, low-cost, and modular testbed designed to support the proposed LLM-enhanced RL framework and to enable Sim2Real zero-shot transfer. The hull integrates 3D-printed ABS components with aluminum structural elements, providing a balance between mechanical durability, corrosion resistance, and rapid prototyping capability. With a total cost of approximately \$1000 USD, the platform is significantly more affordable than conventional UUV systems, while its modular architecture allows for flexible integration of sensing, computation, and actuation payloads. The propulsion system consists of eight custom-built thrusters with thrust characteristics comparable to the Blue Robotics T200, arranged in a fully actuated six-degree-of-freedom configuration to enable independent control of forces and moments along all axes. Vibration isolation is incorporated to mitigate mechanical disturbances and reduce sensor noise.

\setcounter{figure}{3}
\begin{figure}[!t]
    \centering
    \includegraphics[width=0.99\linewidth]{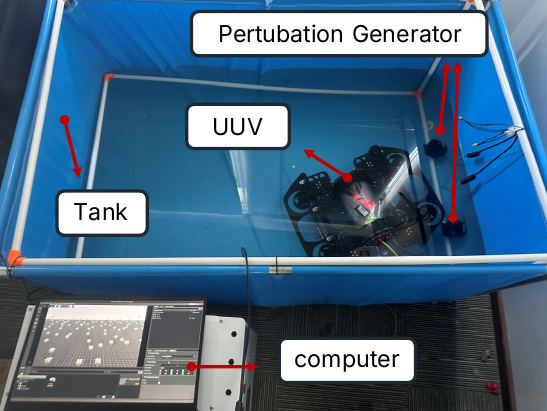}
    \vspace{-5mm}
    \caption{Experimental testbed for real-world validation of EasyUUV.} %
    \label{fig_4}
    \vspace{-3mm}
\end{figure}

\setcounter{figure}{5}
\begin{figure*}[!t]
    \centering
    \includegraphics[width=0.99\linewidth]{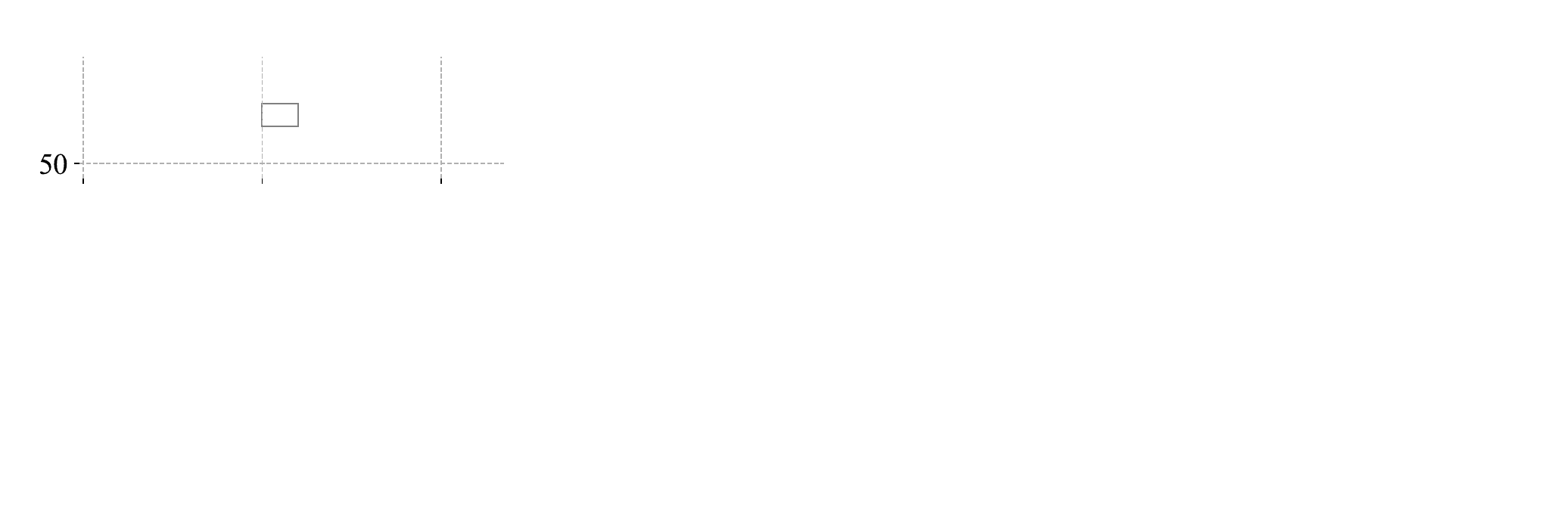}
    \vspace{-2mm}
    \caption{Comparison of UUV attitude tracking and compound error curves for different control strategies in simulation experiments. (Left): attitude tracking response with RL under different controllers. (Middle): attitude tracking response under w/o RL and w/ RL settings. (Right, top): compound error with RL under different controllers; (Right, bottom): compound error under w/o RL and w/ RL settings.}
    \label{fig_7}
    \vspace{-2mm}
\end{figure*}

Low-level control is executed at $100\,\text{Hz}$ by an onboard ESP32-WROOM microcontroller running the pre-calibrated A-S-surface controller, which demands minimal computational overhead (angle-error computation, S-surface mapping, adaptive compensation, thrust allocation, and PWM generation). High-level decision-making is handled by an RL policy on a surface laptop. Real-time communication uses a half-duplex RS-485 tether with a strict time-division slot allocation protocol at $115,200\,\text{bps}$. Combined with rapid RL inference ($<2\,\text{ms}$) and ESP32 sensor acquisition ($<0.5\,\text{ms}$), the closed-loop execution easily converges within the $10\,\text{ms}$ control budget ($100\,\text{Hz}$), leaving a safety margin of $>1\,\text{ms}$ against timing jitter. In each loop, the ESP32 estimates Euler angles from a 9-DOF MPU9250 IMU using robust complementary filtering and transmits them to the host, which converts states to quaternions and executes the PPO policy (a lightweight 9D-to-4D forward mapping that is computationally trivial). The ESP32 then calculates per-axis S-surface outputs $\{u_{\phi},u_{\theta},u_{\psi},u_{d}\}$, distributes them to 8 thrusters via the 4-vertical/4-horizontal layout shown in Fig.~\ref{fig_3}, and maps them to physical PWM signals via the thrust curve of Eq.~(1).

The event-triggered LLM tuning module runs asynchronously on the host computer. In our experiments, the end-to-end LLM response latency averages $5.6\,\text{s}$ ($7.3\,\text{s}$ maximum), including visual encoding, API transmission, inference, and output parsing. This delay only affects the availability of parameter updates and does not impact the deterministic $100\,\text{Hz}$ low-level loop. The host computer records tracking data, computes historical MSE, and compiles visual logs for tuning. Once the LLM returns, the host parses the bounded scaling factors and transmits gain updates to the ESP32; while waiting, the microcontroller continues execution using the latest valid parameter cache. Key real-time parameters governing the system include the low-level control and IMU feedback rates, communication latency, and LLM triggering/response times. Through this tight hardware--software integration, the physical UUV, ESP32, RL policy, and asynchronous LLM are closed into a portable, cost-effective deployment pipeline. The entire system is housed in a 30~L waterproof enclosure, weighs $<20$~kg, and supports single-person deployment and operation, making it both portable and cost-effective for research-grade attitude control experiments. For conciseness, a practical, step-by-step engineering roadmap detailing the cross-platform migration of EasyUUV is provided in Section III of the supplementary material. By closely matching the simulated dynamics and leveraging LLM-based online fine-tuning during deployment, EasyUUV provides a practical and reliable platform for evaluating Sim2Real zero-shot transfer in real-world underwater robotic systems.

\vspace{-2mm}
\section{Experiments}
In this section, we systematically evaluate EasyUUV through simulation and real-world experiments. As shown in Fig. 4, the EasyUUV testbed consists of UUV hardware connected to a host computer, which performs RL training, policy deployment, and real-time sensor data collection. To mimic realistic underwater dynamics, we also apply two dedicated perturbation generators in a confined indoor tank.

\vspace{-3mm}
\subsection{Simulation Setup}
The simulation training was conducted on a computer equipped with a Ryzen 9 7945HX CPU and an RTX 4060 GPU. A total of 460 episodes ($\sim3\times10^7$ steps) were completed in approximately 130 seconds, demonstrating high computational efficiency and rapid policy iteration. It should be noted that this training cost is incurred only offline before deployment. During real-world operation, the online computational load mainly comes from lightweight RL policy inference, ESP32-based A-S-Surface control, RS-485 communication, sensor feedback processing, and asynchronous LLM-guided parameter tuning.

Toward the end of training, we introduce two evaluation tasks to assess the Mean Square Error (MSE) performance of different control strategies. \textbf{Task 1} involves tracking a smooth sinusoidal signal, while \textbf{Task 2} requires following a more complex trajectory constructed by summing multiple sine waves with distinct frequencies:
\begin{equation}
\mathcal{S}(t) = A \cdot \sum_{f \in \mathcal{F}} \sin(2\pi f t),
\end{equation}
where $A$ is the amplitude (in radians), $\mathcal{F}$ is the set of frequencies (in Hz), and $t$ denotes time (in seconds). The specific parameters for each attitude angle are:
\begin{itemize}
    \item \textbf{Yaw:}  $A$ = 1.35, $\mathcal{F}$ = \{-0.1,\, 0.2,\, 0.4,\, 0.8,\, 1.6,\, -3.2\},
    \item \textbf{Pitch:} $A$ = 1.10, $\mathcal{F}$ = \{-0.1,\, 0.2,\, 0.5,\, -1.0,\, 2.0,\, 3.5\}, 
    \item \textbf{Roll:}  $A$ = 0.95, $\mathcal{F}$ = \{0.15,\, 0.3,\, 0.5,\, -0.9,\, 1.8,\, -3.0\}.
\end{itemize}

To further evaluate tracking accuracy, we define the \textbf{compound error} at time \( t \) as the sum of absolute differences between actual and desired yaw, pitch, and roll angles extracted from the corresponding quaternions:
\begin{equation}
\text{CompoundError}_t = \sum_{i \in \{\phi,\, \varphi,\, \theta\}} |i_t - i_{\text{des},t} |.
\end{equation}

\subsection{Simulation Results}
We first conduct the simulation RL training, as shown in Fig.~\ref{fig_5}. The curves compare three controllers, RL with A-S-Surface, S-Surface, and PID, in terms of cumulative reward and MSE. Here, the controller parameters are primarily adopted from \cite{11} to ensure a fair baseline for comparison. Fig.~\ref{fig_5}(Left) and Fig.~\ref{fig_5} (Middle) compare the cumulative reward and MSE of the proposed RL + A-S-Surface controller against RL + S-Surface and RL + PID baselines. The RL + A-S-Surface configuration converges fastest and achieves the highest final reward with the lowest training MSE, highlighting the clear advantage of the adaptive compensation term $\Delta u$ in accelerating policy search. Furthermore, Fig.~\ref{fig_5}(Right) provides a comparative bar chart under Task 1 and Task 2 with and without the high-level RL policy. Across both tasks, the RL + A-S-Surface controller consistently achieves the lowest MSE ($0.54 \times 10^{-2}$ under Task 1 and $0.50 \times 10^{-2}$ under Task 2), yielding substantial tracking precision improvements over both static non-RL baselines and uncompensated S-surface controllers.

Building on this foundation, we next examine the effect of DR on RL training. Specifically, we present MSE results under different DR levels-None (NDR), Small-scale (SDR), and Large-scale (LDR)-to analyze the impact of physical variability on policy generalization. To evaluate out-of-domain performance, we fixed the UUV mass while varying its volume to create two conditions: density at 0.95$\times$ and 1.05$\times$ the default value ($\approx$ water density), denoted as Pos. buoy and Neg. buoy, respectively. The policies were trained with RL using the A-S-Surface controller. As shown in Table II, policies without DR suffer significant MSE degradation under buoyancy shifts, whereas SDR and LDR reduce performance loss, with SDR achieving better generalization and LDR showing less stability. Thus, these findings suggest that exposing policies to broader physical uncertainties during training improves robustness and cross-domain performance.

\begin{table}
  \centering
  \vspace{-2mm}
  \caption{The MSE Results under Varying Domain Randomization Levels in Task 1 and Task 2.} 

    \begin{tabular}{ccccc}
      \toprule
      \multicolumn{2}{c}{Settings} & \textbf{NDR}& \textbf{SDR} & \textbf{LDR}\\
      \midrule
    \multirow{3}{*}{\textbf{Task 1}} & In domain & 0.0054 & \textbf{0.0051} & 0.0061 \\
    & Pos. buoy & 0.0344 & \textbf{0.0087} & 0.0110 \\
    & Neg. buoy & 0.0339 & \textbf{0.0091} & 0.0092 \\
    \midrule
    \multirow{3}{*}{\textbf{Task 2}} & In domain & 0.0050 & 0.0057 & 0.0061 \\
    & Pos. buoy & 0.0388 & \textbf{0.0066} & 0.0160 \\
    & Neg. buoy & 0.0320 & \textbf{0.0079} & 0.0132 \\
      \bottomrule
  \end{tabular}
  \vspace{-3mm}
\end{table}

Having validated robustness against environmental variability, we then turn to attitude tracking performance. To further evaluate attitude tracking in simulation, Fig.~\ref{fig_7} compares yaw, pitch, and roll responses under Task 2. In Fig.~\ref{fig_7}(Left), RL+A-S-Surface achieves the closest tracking with fast convergence and minimal steady-state error, RL+S-Surface shows larger deviations and mild oscillations, while RL+PID responds more slowly with significant errors, especially in pitch and roll. Fig.~\ref{fig_7}(Middle) further shows that A-S-Surface with RL attains higher accuracy and responsiveness than its non-RL counterpart, which exhibits delays and larger errors. Moreover, Fig.~\ref{fig_7}(Right, top) illustrates compound error evolution: RL+A-S-Surface maintains the lowest and most stable error, RL+S-Surface shows moderate fluctuations, and RL+PID suffers larger deviations, particularly around 10-12s and near the end. Finally, Fig.~\ref{fig_7}(Right, bottom) confirms that RL reduces both average error (from $\mu\!\!=\!\!0.452$ to $\mu\!\!=\!\!0.103$) and variability. Overall, the results highlight the advantage of integrating RL with adaptive control for robust multi-axis attitude regulation.

\vspace{-3mm}
\subsection{Environmental Disturbance Robustness}
To evaluate the passive disturbance rejection of EasyUUV under non-stationary fluid environments, we simulate wind-generated ocean waves using the JONSWAP spectrum density and Airy linear wave theory (For conciseness, the complete analytical derivations of the depth-dependent current velocities are provided in Section~II of the supplementary material)~\cite{JONSWAP}.

We evaluate the tracking MSE and control efficiency across six scenarios under Task 2: quiet water (\texttt{none}), steady current (\texttt{constant}, $v_x = 0.25\,\text{m/s}$), sinusoidal current (\texttt{sine}), mild waves (\texttt{jonswap\_mild}, $H_s = 1.0\,\text{m}$), severe storm waves (\texttt{jonswap\_strong}, $H_s = 2.5\,\text{m}$), and a composite wave-current field (\texttt{current\_plus\_jonswap}, $0.359\,\text{m/s}$ steady flow with severe waves). The global tracking metrics over a 200-second simulation are summarized in Table~\ref{tab:disturbance_results}.

\begin{table}[!t]
\centering
\vspace{-2mm}
\setlength{\tabcolsep}{4.5pt} 
\caption{UUV Tracking Accuracy under Oceanographic Disturbances (Task 2)}
\label{tab:disturbance_results}
\begin{tabular}{l|c|c}
\hline
\textbf{Scenario} & \textbf{Mean MSE} ($\text{rad}^2$) & \textbf{Max MSE} ($\text{rad}^2$) \\ \hline
\texttt{none} & 0.0054 & 0.0281 \\
\texttt{constant} & 0.0049 & 0.0306  \\
\texttt{sine} & 0.0059 & 0.0288  \\
\texttt{jonswap\_mild} & 0.0050 & 0.0322  \\
\texttt{jonswap\_strong} & 0.0048 & 0.0285  \\
\texttt{current\_plus\_jonswap} & 0.0048 & 0.0353 \\ \hline
\end{tabular}
\vspace{-2mm}
\end{table}

Under stochastic JONSWAP waves, the Mean MSE remains comparable to, and in some cases slightly lower than, the undisturbed baseline, demonstrating the strong passive disturbance rejection capability of EasyUUV. Even under the more challenging \texttt{current\_plus\_jonswap} scenario, the Mean MSE is maintained at 0.0048 with only a moderate increase in Max MSE, indicating that EasyUUV can preserve accurate and stable tracking under composite wave-current forcing.

\subsection{Cross-Embodiment Zero-Shot Transfer}
The evaluation of generalization capability across structurally different platforms is conducted through zero-shot evaluations across four distinct UUV configurations under Task 2 in simulation: the default \texttt{Base}, a torpedo-shaped AUV (\texttt{Long Body}, with $2.5\times$ pitch/yaw inertia), a heavier ROV (\texttt{Heavy Duty}, with $2\times$ mass and $5\times$ drag), and an asymmetric payload variant (\texttt{Asym.}, with $5\,\text{cm}$ lateral COM offset). The complete physical specification sheets are provided in Section~III of the supplementary material. The zero-shot tracking metrics without retraining policies are summarized in Table~\ref{tab:embodiment_results}.

\begin{table}[!t]
\centering
\vspace{-2mm}
\setlength{\tabcolsep}{5pt} 
\caption{Generalization Metrics Across Embodiments (Task 2)}
\label{tab:embodiment_results}
\begin{tabular}{l|c|c}
\hline
\textbf{Embodiment} & \textbf{Controller Gain Scale} & \textbf{Mean MSE} ($\text{rad}^2$) \\ \hline
\texttt{Base} & Maintain $1\times$ Gain & 0.0054 \\
\texttt{Long Body} & Decrease $\zeta_{1}$, Increase $\zeta_{2}$ & 0.0154 \\
\texttt{Heavy Duty} & Increase $\zeta$ to $3\times$ & 0.0157 \\
\texttt{Asym.} & Increase $\zeta_1$ to $3\times$ & 0.0068 \\ \hline
\end{tabular}
\vspace{-2mm}
\end{table}

\begin{table}[!t]
\centering
\vspace{-2mm}
\setlength{\tabcolsep}{5pt} 
\caption{Tracking MSE Comparison across Fault Scenarios}
\label{tab:fault_wave_performance_metrics}
\begin{tabular}{l|c|c}
\hline
\multirow{2}{*}{\textbf{Control Mode}} & \textbf{Mean MSE} & \textbf{Mean MSE} \\
 & (\texttt{trend\_fault}) & (\texttt{wave\_plus\_fault}) \\ \hline
Pure RL (Static) & 0.0104 & 0.0094 \\
Fuzzy (Constrained) & 0.0096 & 0.0105 \\
LLM & \textbf{0.0053} & \textbf{0.0049} \\ \hline
\end{tabular}
\vspace{-2mm}
\end{table}

The quantitative metrics in Table~\ref{tab:embodiment_results} substantiate the platform-agnostic control capability of the proposed architecture. When the physical properties vary drastically, the closed-loop system maintains high tracking stability and bounded tracking errors. For the heavy-class \texttt{Heavy Duty} variant, although extreme hydrodynamic drag and mass variations degrade the static policy's performance, stability is preserved up to the hard physical thrust limits. This zero-shot capability establishes that the underlying policy can generalize across platforms.

\subsection{Active LLM-Based Adaptive Tuning}
When physical degradation or environmental perturbations exceed the passive robustness limits of the static policy, we activate our LLM-based active adaptive tuning loop. In this experiment, the LLM does not perform direct control, online policy learning, or model fine-tuning; instead, it analyzes tracking trends and tuning history, and adjusts selected controller parameters using predefined scaling rules.

We inject an unobservable progressive fault into the two rear-mounted yaw-governing thrusters (linear decay rate $r = 0.02$/s starting at $t=20$~s, reaching complete failure at $t=70$~s). As a representative classical self-tuning baseline, we implement a Mamdani-type Fuzzy Logic Controller (\texttt{Fuzzy}), which outputs a unified scaling multiplier applied to all control loops. Full experimental setups are detailed in Section~V of the supplementary material. The tracking metrics under the pure fault (\texttt{trend\_fault}) and wave-current composite perturbations (\texttt{wave\_plus\_fault}) are summarized in Table~\ref{tab:fault_wave_performance_metrics}.

\begin{figure}[!t]
    \centering
    \includegraphics[width=0.91\linewidth]{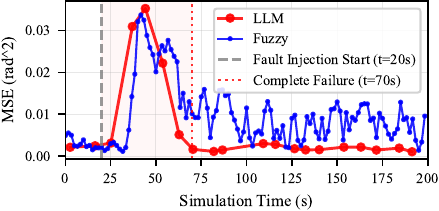}
    \vspace{-3mm}
    \caption{Per-decision tracking MSE comparison between the LLM and the improved Fuzzy Controller (\texttt{trend\_fault}).}
    \label{fig:llm_mse}
    \vspace{-2mm}
\end{figure}

\begin{figure}[!t]
    \centering
    \includegraphics[width=1.00\linewidth]{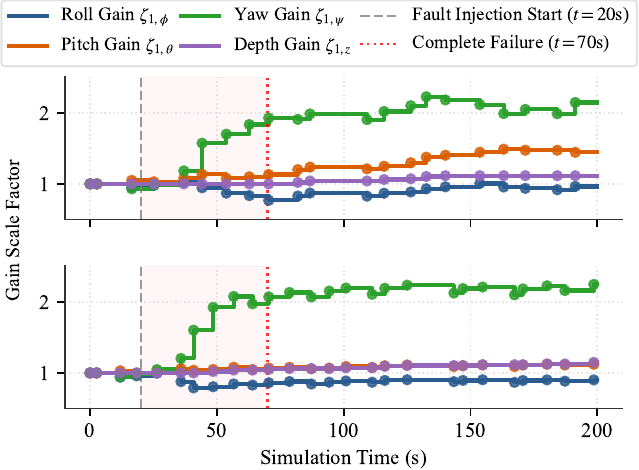}
    \vspace{-7mm}
    \caption{Comparative analysis of decoupled proportional gain trajectories ($\zeta_1$) adjusted by the LLM over the 200\,s evaluation timeline under: (a) the pure thruster fault scenario (\texttt{trend\_fault}), and (b) the composite wave-current forcing plus thruster fault scenario (\texttt{wave\_plus\_fault}).}
    \label{fig:llm_comparative_gains}
    \vspace{-2mm}
\end{figure}

\begin{figure*}[!t]
    \centering
    \includegraphics[width=0.99\linewidth]{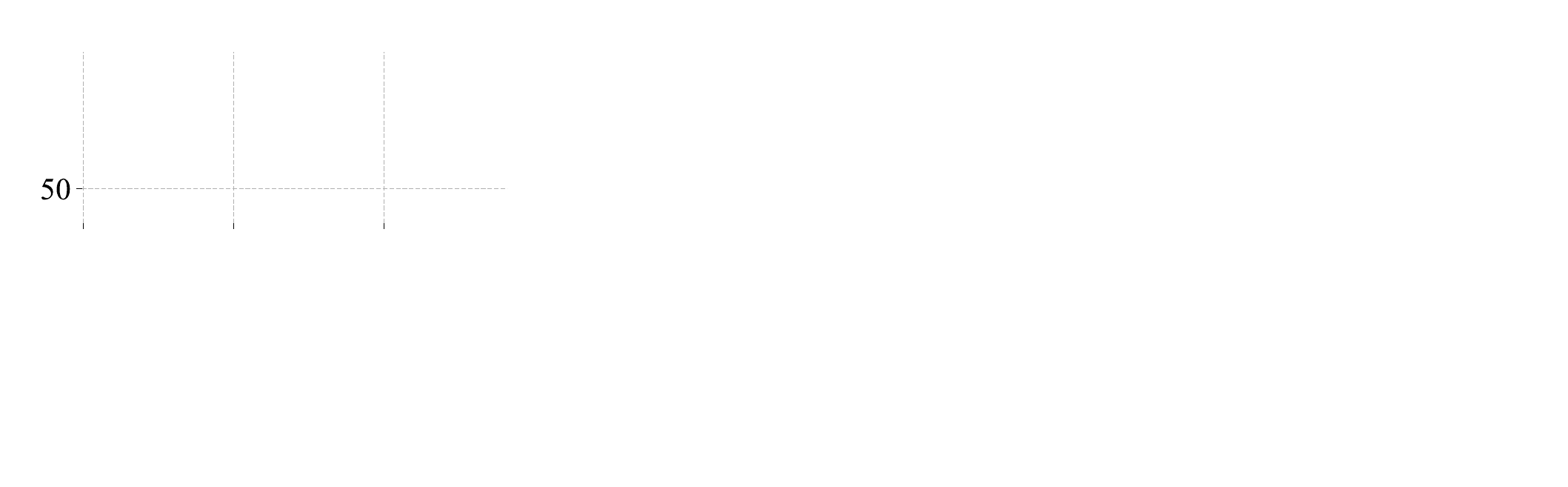}
    \vspace{-2mm}
    \caption{Comparison of UUV attitude tracking and compound error curves for different attitude angle combinations, under both RL and non-RL settings in real-world experiments. (Left): tracking response of Yaw and Roll. (Middle): tracking response of Yaw and Pitch. (Right, top): compound error of Yaw and Roll; (Right, bottom): compound error of Yaw and Pitch.}
    \label{fig_9}
    \vspace{-2mm}
\end{figure*}

\begin{figure*}[!t]
    \centering
    \includegraphics[width=0.99\linewidth]{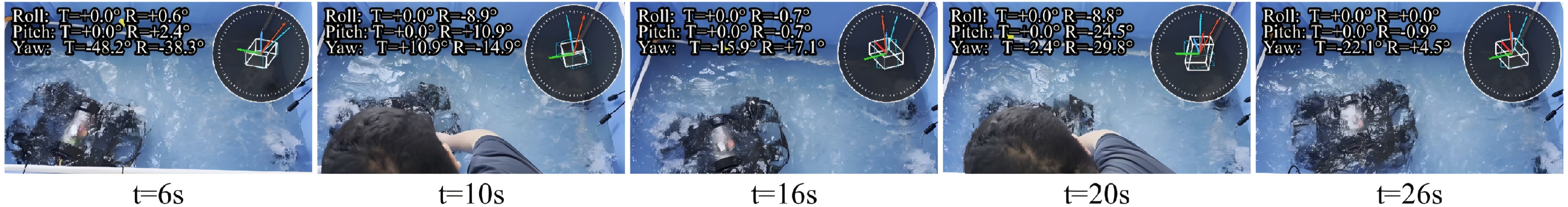}
    \vspace{-2mm}
    \caption{Time-synchronized 5-frame snapshot sequence of EasyUUV operating in an indoor tank under turbulent perturbations ($t = 6\text{ s}, 10\text{ s}, 16\text{ s}, 20\text{ s}, 26\text{ s}$). The aligned frames are annotated with top-left numerical states (T: Target, R: Real) and top-right 3D-head-up display overlays to demonstrate stable attitude tracking and regulation in the presence of dynamic and non-reproducible water disturbances.}
    \label{fig_11}
    \vspace{-2mm}
\end{figure*}

\begin{figure*}[!t]
    \centering
    \includegraphics[width=0.99\linewidth]{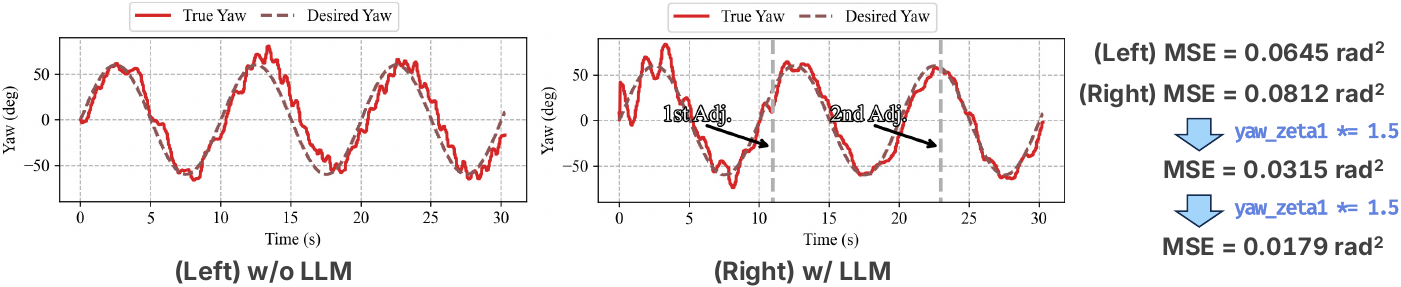}
    \vspace{-2mm}
    \caption{Tracking response curves under turbulent perturbations with and without LLM-based online fine-tuning of controller parameters. }
    \label{fig_12}
    \vspace{-2mm}
\end{figure*}

As shown in Table~\ref{tab:fault_wave_performance_metrics}, the LLM agent achieves the lowest overall Mean MSE ($0.0049\ \text{rad}^2$ under \texttt{wave\_plus\_fault}), outperforming the robust Fuzzy baseline by up to $53.6\%$. Furthermore, Fig.~\ref{fig:llm_mse} displays the tracking MSE, while Fig.~\ref{fig:llm_comparative_gains} depicts the decoupled proportional gain trajectories ($\zeta_1$) adjusted by the LLM over the 200-second timeline.

Due to the average $5.6$~s API response latency, the LLM's corrective adjustments lag slightly, causing the tracking error to peak at $t = 44.2$~s with an MSE of $0.0353\ \text{rad}^2$ in Fig.~\ref{fig:llm_mse}. However, upon receiving the visual log, the LLM identifies the asymmetric thruster decay and executes an aggressive yaw gain decrease, dropping the error within two decision steps and converging to $0.0020\ \text{rad}^2$ at steady-state. Unlike \texttt{Fuzzy}'s unified scalar tuning, the LLM optimizes gains on a systemic, multi-axis decoupled level (the complete 200-step decision history logs and system prompts are disclosed in Section~IV of the supplementary material).


\setcounter{figure}{12}
\begin{figure*}[!t]
    \centering
    \includegraphics[width=0.99\linewidth]{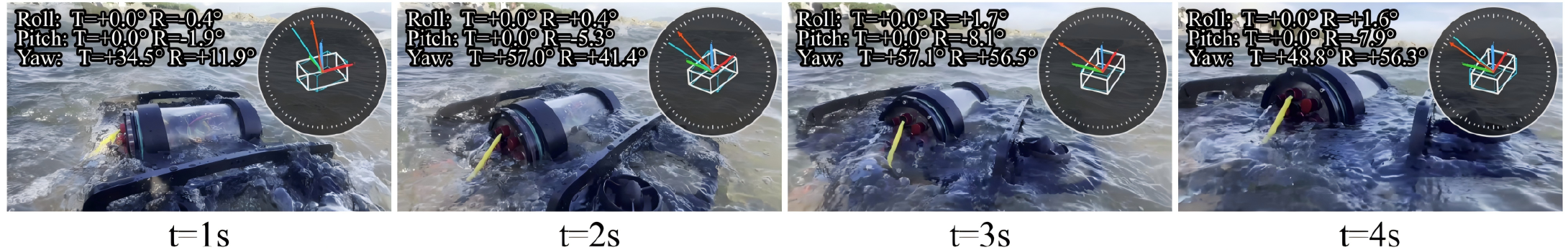}
    \vspace{-2.5mm}
    \caption{Time-synchronized 4-frame snapshot sequence of EasyUUV operating under open-ocean wave disturbances during sea trials ($t = 1\text{ s}, 2\text{ s}, 3\text{ s}, 4\text{ s}$). The aligned frames are annotated with top-left numerical states (T: Target, R: Real) and top-right 3D-head-up display overlays to demonstrate stable attitude tracking and regulation under dynamic and non-reproducible marine conditions.}
    \label{fig_14}
    \vspace{-2mm}
\end{figure*}

\setcounter{figure}{11}
\begin{figure}[!t]
    \centering
    \includegraphics[width=0.99\linewidth]{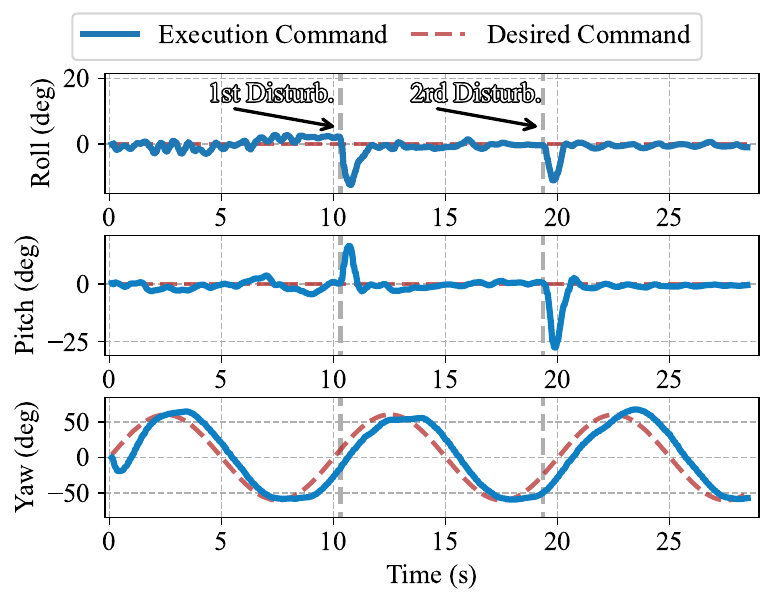}
    \vspace{-6.5mm}
    \caption{Tracking response curves along the roll, pitch, and yaw axes in tank experiments under turbulent and transient perturbation.}
    \label{fig_13}
    \vspace{-2mm}
\end{figure}

\vspace{-2mm}
\subsection{Real-World Deployment}
Building on these comprehensive simulation and adaptive tuning evaluations, we transition to real-world validation on the physical UUV platform. We first conduct the tank experiment under disturbance-free conditions to evaluate EasyUUV's Sim2Real zero-shot transfer capability using the expert-level RL policy directly taken from simulation with the A-S-Surface controller. As shown in Fig.~\ref{fig_9}(Left) and Fig.~\ref{fig_9}(Middle), the RL-enabled controller tracks desired commands more closely, keeping roll and pitch near the origin with reduced drift and phase lag, while the non-RL case shows larger deviations. In addition, the right-hand panels of Fig.~\ref{fig_9} further compare compound error curves, where the RL-enabled controller achieves lower average error ($\mu\!\!=\!\!0.2356$\! vs. $0.3836$, and $\mu\!\!=\!\!0.2421$ vs. $0.2876$) and reduced variability ($\sigma\!\!=\!\!0.080$ vs. $0.150$, and $\sigma\!\!=\!\!0.0896$ vs. $0.0970$), indicating improved robustness. Taken together, these findings preliminarily confirm that EasyUUV can achieve effective zero-shot transfer from simulation to real-world deployment, significantly enhancing multi-axis tracking performance.

After the initial disturbance-free tests, we activated the perturbation generators to evaluate whether the LLM could improve bounded runtime parameter adaptation under disturbances. To quantify the contribution of the multimodal LLM, we compared the yaw tracking performance before and after LLM-guided parameter adaptation under the same turbulent perturbation setting. As shown in Fig.~10, which presents a time-synchronized 5-frame snapshot sequence with unified annotations, top-left numerical attitude states, and top-right 3D head-up display overlays, EasyUUV rapidly suppresses disturbances and restores the vehicle to the desired trajectory. Fig.~\ref{fig_12} further shows that the yaw tracking MSE decreases from $0.0812\ \mathrm{rad}^2$ to $0.0179\ \mathrm{rad}^2$ after two LLM-guided adjustments, corresponding to an absolute reduction of $0.0633\ \mathrm{rad}^2$ and an approximate relative reduction of $78.0\%$. Since this improvement is achieved without retraining the RL policy, fine-tuning the LLM, or conducting online trial-and-error search, it directly reflects the contribution of the constrained LLM-based runtime parameter adaptation module. These results support the feasibility of multimodal LLM-guided parameter tuning for improving real-world UUV attitude control under disturbances.

\setcounter{figure}{13}
\begin{figure}[!t]
    \centering
    \includegraphics[width=0.99\linewidth]{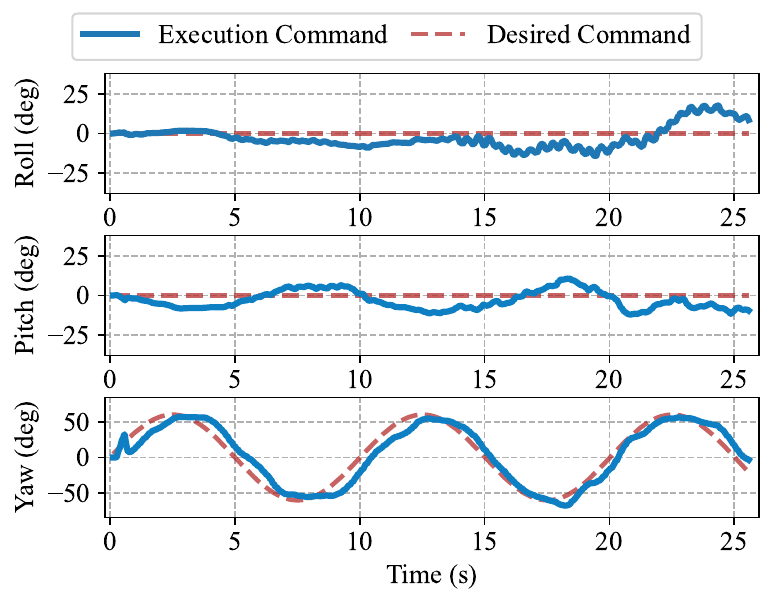}
    \vspace{-6.5mm}
    \caption{Tracking response curves along the roll, pitch, and yaw axes in sea trials under wave-induced turbulence.}
    \vspace{-2mm}
    \label{fig_15}
\end{figure}

To more rigorously assess robustness, two strong transient perturbations are manually introduced at 10.3s and 19.4s. As illustrated in Fig.~\ref{fig_13}, under turbulent disturbances, the EasyUUV still tracks the desired trajectory closely on all three axes with steady-state errors near zero. Although roll and pitch briefly deviate when the manual perturbations occur, the framework quickly suppresses the errors and restores the trajectory, while yaw tracking remains accurate throughout. Together with the enhanced snapshot sequence in Fig.~10, these results verify the robustness and stability of the proposed framework, enabling EasyUUV to maintain high-precision control under turbulent and sudden disturbances while demonstrating strong Sim2Real transfer capability.

Based on the above tests, we finally extend the evaluation to sea trials. Fig.~13 and Fig.~\ref{fig_15} present the results, complementing the earlier tank experiments and highlighting the framework's zero-shot domain transfer capability. The tracking curves indicate that EasyUUV closely follows the desired commands in roll, pitch, and yaw under wave-induced turbulence, with steady-state errors near zero. Specifically, Fig.~13 presents a time-synchronized 4-frame snapshot sequence at $t=1\text{ s}$, $2\text{ s}$, $3\text{ s}$, and $4\text{ s}$, with unified visual annotations, numerical attitude states, and 3D head-up display overlays to more clearly illustrate the transient yaw and pitch adjustments under real ocean conditions. Together with the tank experiments, these results confirm that the framework transfers directly from controlled environments to open-sea settings without retraining, achieving robust disturbance rejection and stable high-precision control.

\vspace{-3mm}
\section{Conclusions}
\vspace{-0.8mm}
In this paper, we introduce EasyUUV, an LLM-enhanced platform-agnostic and lightweight Sim2Real RL framework for robust UUV attitude control. The framework integrates domain-randomized RL training with a hybrid control architecture that incorporates an A-S-Surface controller, while a multimodal LLM agent provides runtime parameter fine-tuning without additional retraining. Built on a cost-effective 6-DoF UUV platform, EasyUUV enables efficient simulation-based policy learning and achieves zero-shot transfer to real-world deployment. Extensive simulation and field experiments demonstrate that EasyUUV offers stable, generalizable control, with robust tracking performance and consistent Sim2Real performance under diverse underwater conditions.

\textbf{Limitations and Future Work:} While EasyUUV demonstrates promising Sim2Real robustness and generalization, several limitations remain. First, the underwater Sim2Real gap remains challenging, as hydrodynamic forces, buoyancy variations, thruster non-linearities, and sensor noise are tightly coupled and difficult to model completely. Although domain randomization improves robustness to parameter variations, the current simulator cannot fully capture long-term hydrodynamic degradation, severe turbulence, or highly uncertain disturbances. Second, we must still balance learning-based adaptability with control stability. To mitigate safety risks, the RL policy only outputs high-level correction commands, while the local S-surface controller performs structured low-level execution. The RL policy is trained via parallelized simulation rather than pre-collected real-world datasets, where domain randomization helps prevent overfitting to a single nominal model. Meanwhile, the LLM-based tuning module provides training-free, bounded runtime adaptation based on tracking trends, visual logs, and textual feedback. While these designs support generalization beyond the original simulation distribution, we do not claim universal applicability to all UUV platforms or autonomy tasks. Formal stability guarantees for the LLM-guided tuning are still lacking, and our current physical validation remains focused on a single lightweight platform and attitude tracking tasks. Future work will focus on improving simulation fidelity, developing safety-supervised and stability-aware LLM tuning, and executing long-term field trials across diverse commercial UUV platforms and autonomy tasks, including trajectory tracking, navigation, and inspection.


\bibliographystyle{Bibliography/IEEEtranTIE}
\bibliography{Bibliography/IEEEexample}\ 

\begin{IEEEbiography}[{\includegraphics[width=1in,height=1.25in,clip,keepaspectratio]{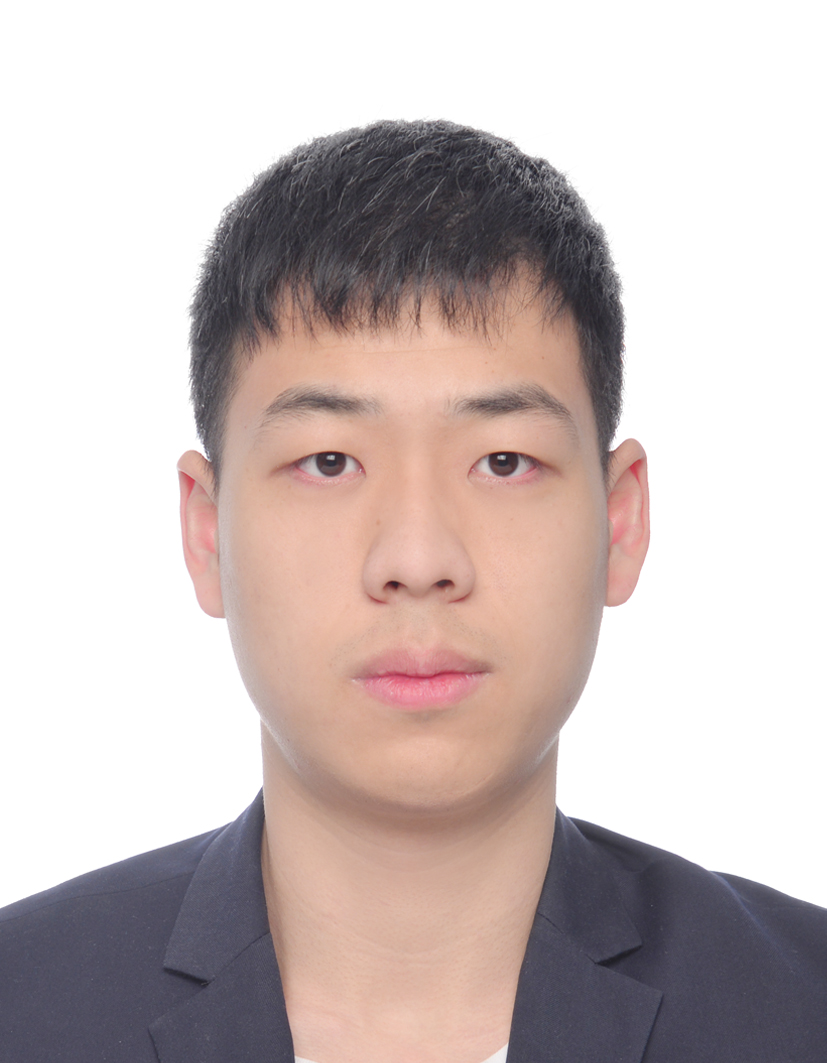}}]{Guanwen Xie} (Student Member, IEEE) received his B.E. degree in Ocean Engineering and Technology at Ocean College from Zhejiang University, and he is currently pursuing the M.S. degree in Electronic and Information Engineering from Tsinghua Shenzhen International Graduate School, Tsinghua University, China. His main research interest is applying reinforcement learning and large language models in underwater intelligent decision-making. Besides, he is also an outstanding undergraduate of Zhejiang University.
\vspace{-6mm}
\end{IEEEbiography}

\begin{IEEEbiography}
[{\includegraphics[width=1in,height=1.25in,clip,keepaspectratio]{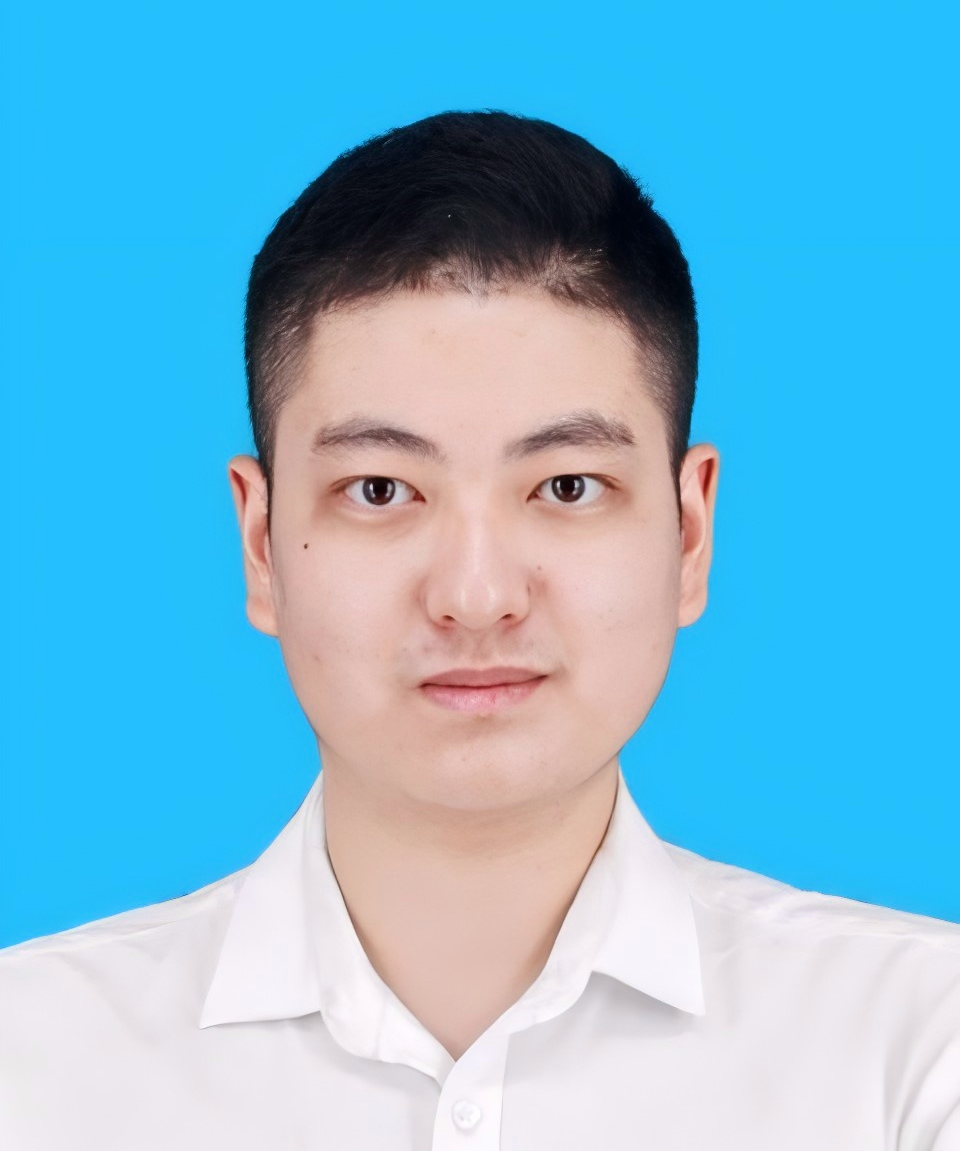}}]{Jingzehua Xu} (Student Member, IEEE)
received his B.S. degree in Marine Science from Zhejiang University, Hangzhou, China in 2023, and received his M.S. degree in Electronic and Information Engineering from Tsinghua Shenzhen International Graduate School, Tsinghua University, China in 2025. His main research interests include reinforcement learning, physical oceanography, and their applications in underwater scenarios. Besides, he is the outstanding undergraduate at Zhejiang University. 
\vspace{-6mm}
\end{IEEEbiography}

\begin{IEEEbiography}
[{\includegraphics[width=1in,height=1.25in,clip,keepaspectratio]{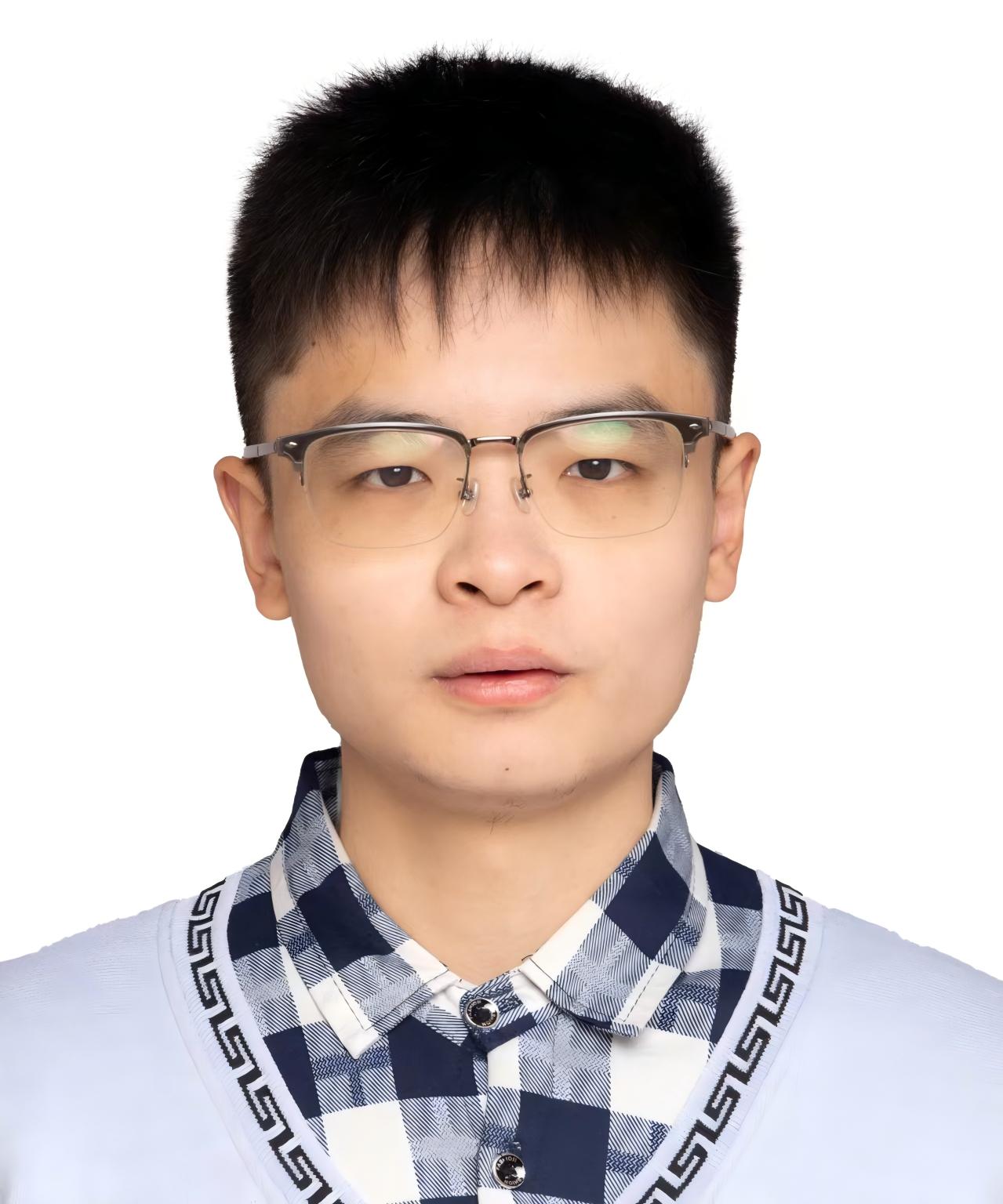}}]{Jiwei Tang} (Student Member, IEEE) obtained his Bachelor's degree in Marine Engineering and Technology from Zhejiang University, Hangzhou, China in 2022, and Master's degree in Electrical and Computer Engineering from National University of Singapore, Singapore in 2024. He is currently pursuing his Ph.D. in the Department of Data and Systems Engineering at The University of Hong Kong. His research interests include nonlinear system control, constrained control, deep reinforcement learning, and autonomous systems.
\vspace{-6mm}
\end{IEEEbiography}

\begin{IEEEbiography}
[{\includegraphics[width=1in,height=1.25in,clip,keepaspectratio]{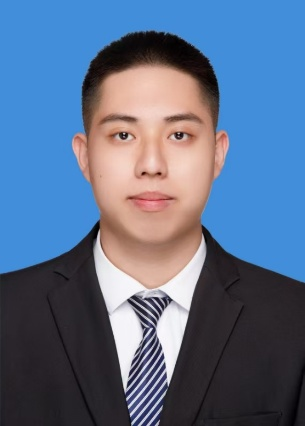}}]{Yubo Huang} is a researcher in Key Laboratory of Transportation Tunnel Engineering, Ministry of Education, Southwest Jiaotong University. His main research is visual representation, domain adaptation, reinforcement learning and AI for science.
\vspace{-6mm}
\end{IEEEbiography}

\begin{IEEEbiography}
[{\includegraphics[width=1in,height=1.25in,clip,keepaspectratio]{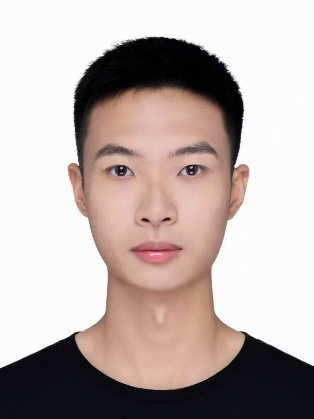}}]{Zixi Wang} is a M.S. student in Software Engineering, University of Electronic Science and Technology of China. His main research is vision language model, domain adaptation, reinforcement learning and AI for science.
\vspace{-6mm}
\end{IEEEbiography}

\begin{IEEEbiography}[{\includegraphics[width=1in,height=1.25in,clip,keepaspectratio]{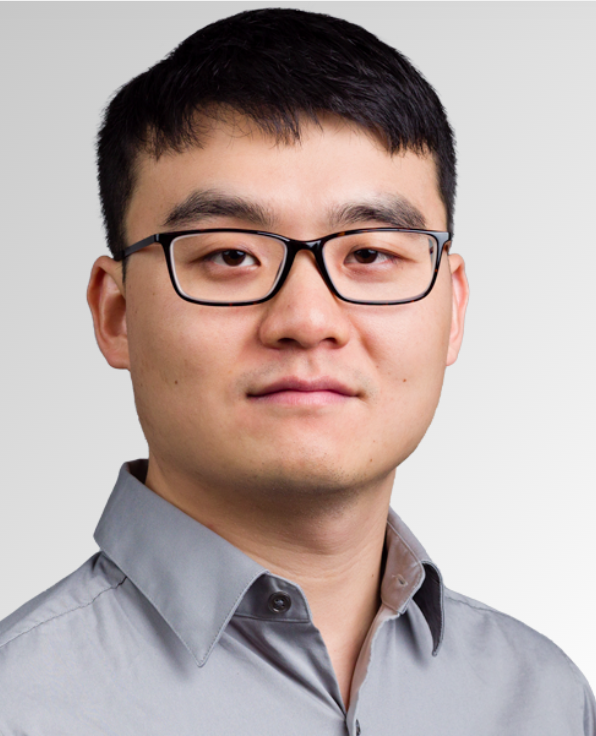}}]{Shuai Zhang} 
(Member, IEEE) received his B.E. degree from the University of Science and Technology of China, Hefei, China, in 2016, and his Ph.D. degree from Rensselaer Polytechnic Institute, Troy, NY, USA, in 2021. From 2022 to 2023, he was a Postdoctoral Research Associate at Rensselaer Polytechnic Institute. He is currently an Assistant Professor in the Department of Data Science at the New Jersey Institute of Technology, Newark, NJ, USA. His research focuses on the theoretical foundations of deep learning and the development of principled, efficient algorithms to improve the reliability and performance of AI applications. He has served as a reviewer or program committee member for NeurIPS, ICML, AAAI, ICLR, AISTATS, TMLR, IEEE TSP, IEEE TNNLS, and IEEE TIT.
\vspace{-6mm}
\end{IEEEbiography}

\begin{IEEEbiography}[{\includegraphics[width=1in,height=1.25in,clip,keepaspectratio]{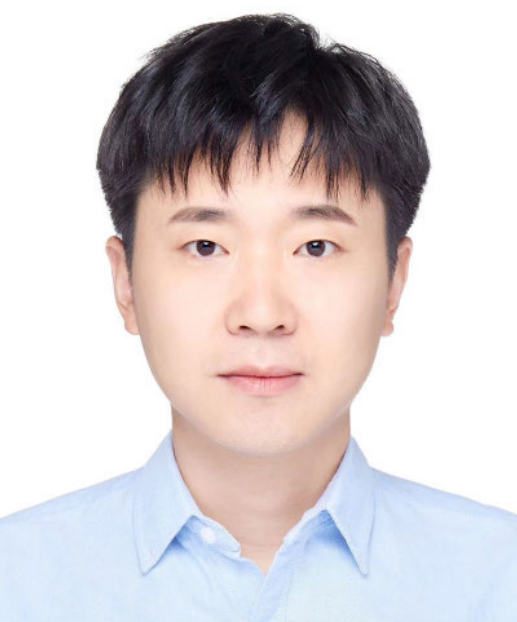}}]{\textbf{Dongfang Ma}} (Member, IEEE) received the M.S. and Ph.D. degrees in information engineering and control from Jilin University, Changchun, China, in 2009 and 2012, respectively. From 2012 to 2014, he was a Post-Doctoral Researcher with the Department of Civil Engineering and Architecture, Zhejiang University, where he has been with the Institute of Marine Sensing and Networking since 2015. Since 2019, he has also been a part-time Research Fellow with the Peng Cheng Laboratory. His current research interests include transportation big data mining and intelligent transportation systems.
\vspace{-6mm}
\end{IEEEbiography}

\begin{IEEEbiography}[{\includegraphics[width=1in,height=1.25in,clip,keepaspectratio]{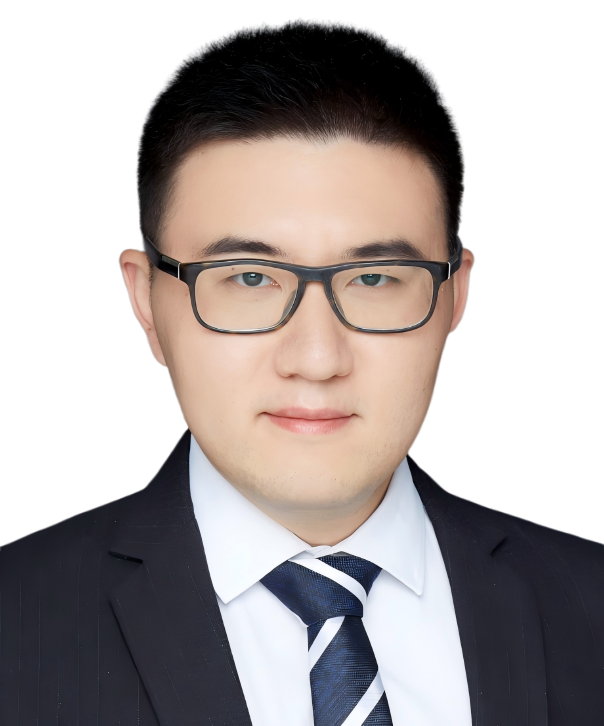}}]{\textbf{Juntian Qu}} (Member, IEEE) received the Ph.D. de gree in mechanical engineering from McGill Univer sity, Montreal, QC, Canada, in 2019. He was the Visiting Ph.D. and Research Assistant with the University of Toronto, from 2017 to 2019. He was the Shuimu Postdoctoral Fellow and Assistant Research Fellow with the Department of Mechanical Engineering, Tsinghua University, Beijing, China, from 2019 to 2021. He is currently an Associate Professor with Shenzhen International Graduate School, Tsinghua University. He has authored or coauthored over 70 academic papers and was granted over 40 patents. His current research interests include underwater soft robot, soft gripper, and advanced sensing technologies.
\vspace{-6mm}
\end{IEEEbiography} 

\begin{IEEEbiography}[{\includegraphics[width=1in,height=1.25in,clip,keepaspectratio]{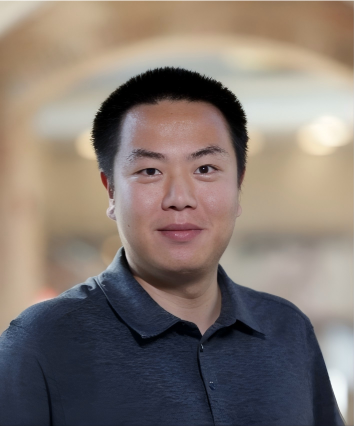}}]
{\textbf{Xiaofan Li}} is currently an Assistant Professor of Mechanical Engineering at the University of Hong Kong. He obtained his B.Sc. from Xi’an Jiaotong University in 2015 and his Ph.D. from Virginia Tech in 2020, both in Mechanical Engineering. Following his doctoral studies, he pursued postdoctoral training at Virginia Tech and the University of Michigan. Prior to joining HKU, he served as an Assistant Research Scientist in the Department of Naval Architecture and Marine Engineering at the University of Michigan. His current research focuses on the dynamics and control of marine systems, with particular emphasis on marine renewable energy and marine robotics.
\vspace{-6mm}
\end{IEEEbiography}

\end{document}